\documentclass[a4paper]{article}

\usepackage{INTERSPEECH2022}
\usepackage{todonotes}

\usepackage{colortbl}
\usepackage{geometry}
\usepackage{booktabs}

\title{Qwen vs. Gemma Integration with Whisper: A Comparative Study in Multilingual SpeechLLM Systems}
\name{Tuan Nguyen$^{*1}$, Long-Vu Hoang$^{*1,2}$, Huy-Dat Tran$^1$}
\address{
  $^1$Institute for Infocomm Research (I$^{2}$R), A$^{\ast}$STAR, Singapore\\
  $^2$SoICT, Hanoi University of Science and Technology, Vietnam}
\email{\{tuan\_nguyen, hdtran\}@i2r.a-star.edu.sg, longvu200502@gmail.com}

\begin{document}

\maketitle
\begin{abstract}
  This paper presents our system for the MLC-SLM Challenge 2025, focusing on multilingual speech recognition and language modeling with large language models (LLMs). Our approach combines a fine-tuned Whisper-large-v3 encoder with efficient projector architectures and various decoder configurations. We employ a three-stage training methodology that progressively optimizes the encoder, projector, and LLM components. Our system achieves competitive performance with a private test average WER/CER result of 16.63\% using the Gemma3-12B and 18.6\% using the Qwen2.5-7B as decoder-only language model.\let\thefootnote\relax\footnotetext{$^*$Equal contribution. Work done during Long Vu's internship at A$^*$STAR, Singapore.}
\end{abstract}
\noindent\textbf{Index Terms}: multilingual speech recognition, speech language modeling, Whisper, transformer architectures

\section{Introduction}
In recent years, Large Language Models (LLMs) have emerged as transformative tools across a wide range of natural language processing (NLP) applications, including machine translation, question answering, summarization, and dialogue systems \cite{li2023blip, liu2023visual, gao2023llama}. Their ability to model long-range dependencies and generate coherent, contextually rich language has made them foundational in both research and industry. As their capabilities continue to evolve, a growing body of work has turned toward leveraging LLMs for speech-related tasks, aiming to unify language and speech processing under a single modeling framework \cite{ma2024embarrassingly, rubenstein2023audiopalm, he2024meralion}. This shift has  opened new directions in Automatic Speech Recognition (ASR), audio captioning, and the development of spoken dialogue systems, particularly in multilingual and real-world settings.

\begin{figure}[h!]
    \centering
    \includegraphics[width=\linewidth]{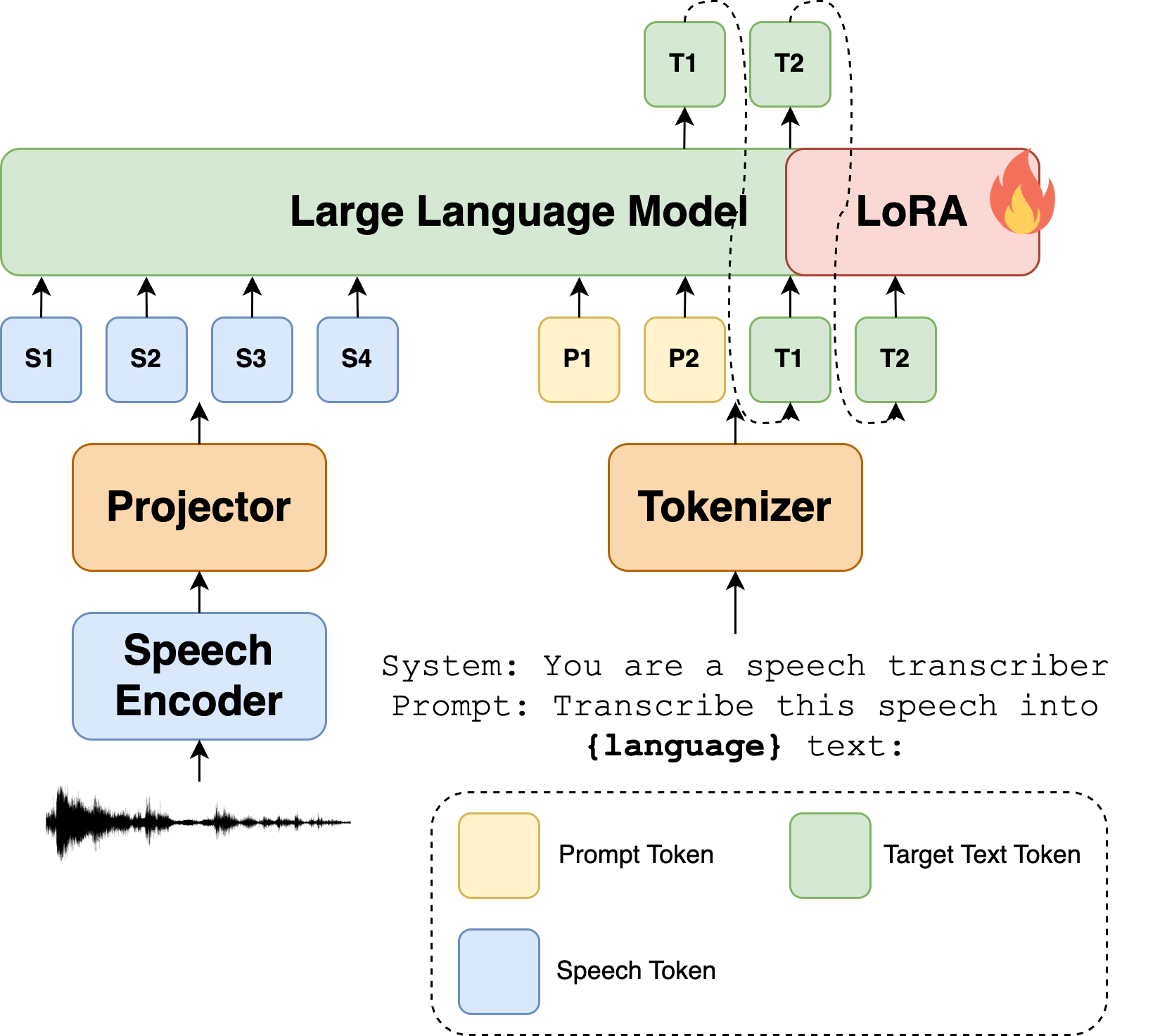}
    \caption{The overall architecture. Main components include a speech encoder, a projector, and a large language model.}
    \label{fig:architecture}
\end{figure}

To address the unique challenges of speech, recent efforts have focused on extending LLMs with speech understanding capabilities through multimodal architectures. These systems typically consist of a speech encoder, a projector module to align modalities, and a language model for decoding. Notable approaches include compressing speech representations temporally, incorporating modality alignment mechanisms, and partially fine-tuning LLMs to adapt to spoken input \cite{ma2024embarrassingly}. Despite such advances, the design of effective LLM-based speech models remains non-trivial, particularly when confronted with real-world conversational speech—characterized by disfluencies, speaker overlaps, and diverse turn-taking styles. Furthermore, the lack of extensive multilingual conversational corpora further complicates generalization and robustness.

In our submission to the MLC-SLM Challenge\footnote{https://www.nexdata.ai/competition/mlc-slm}, we propose a streamlined and effective system architecture that harnesses the strengths of pretrained models with minimal task-specific engineering. Our system utilizes OpenAI’s Whisper model \cite{radford2023robust} as the speech encoder due to its strong generalization capabilities and robustness to multilingual input. For the language modeling component, we explore both Qwen2.5 \cite{team2024qwen2} and Gemma3 \cite{team2025gemma}. A lightweight linear projector module is trained to bridge the speech and language modalities. Through this simple yet effective setup, we demonstrate competitive performance in multilingual conversational speech modeling, highlighting the strength of modular design and pre-trained components over heavily customized architectures\footnote{Implementations at: https://github.com/tuanio/i2r-speechllm-mlc-slm}.


\section{System Architecture}
\label{sec:system}
The architecture of our system is illustrated in Figure \ref{fig:architecture}, including three main components. From the raw waveform $O$, a \textbf{speech encoder} $\text{SE}(\cdot)$ is utilized to extract speech representations from the raw waveform $\tilde{S}=\text{SE}(O) \in \mathbb{R}^{T_s\times D_s}$, where $T_s$ is the number of speech frames and $D_s$ is the output dimension of the speech encoder. Subsequently, the representation is mapped into the same embedding dimension as the LLM's input with a linear transformation, denoted as $S'=\text{Linear}(\tilde{S}) \in \mathbb{R}^{T_s \times D_l}$. After that, the \textbf{projector} learns to compress $S'$ in the temporal dimension and maps them into the text space of the LLM, aligning the different modalities effectively. The projected speech representations is denoted as $S=\text{Projector}(S)\in \mathbb{R}^{T\times D_l}$, where $T < T_s$ is the number of speech time frames after compression by a pooling operation. The compression significantly reduces computational requirements while maintaining essential temporal information needed for the LLM to learn.

The input to the \textbf{LLM} is a concatenation of speech representations $S=(S_t \in \mathbb{R}^{D_l} | t=1,..,T)$ and the instruction tokens $P=(P_n \in \mathbb{R}^{D_l} | n=1,..,N)$, where $N$ is the number of tokens in the instruction. During training, the ground truth transcription are tokenized into token IDs using the LLM's tokenizer. Those token IDs are fed into the LLM as labels and generated via the next-token prediction.

We employ a 3-stage training process for our system. Specifically:
\begin{itemize}
    \vspace{-1mm}
    \item \textbf{Stage 1.} Only the speech encoder is trained
    \item \textbf{Stage 2.} Trained both speech encoder and projector
    \item \textbf{Stage 3.} The projector is trained together with the LoRA adapter in the LLM.
\end{itemize}
\section{Experiment Setup}
\subsection{Models}
\subsubsection{Speech encoder}
We investigate the use of Whisper as a speech encoder, specifically the large-v3 version. Whisper is a Transformer-based encoder-decoder model, trained on 680k hours of labelled speech of multiple languages. The large version has 1.5B parameters.
\subsubsection{Projector}
The projector architecture is a two-layer perceptron with SwiGLU \cite{shazeer2020glu} activation function. There are two projector variants with different compression ratio:
\begin{itemize}
    \item \textbf{Projector 5.} Reduces 1,500 frames to 300 frames in the temporal dimension (1,500 is the number of frames from a 30-second input utterance). This results in a 5:1 compression ratio.
    \item \textbf{Projector 4.} Reduces 1,500 frames to 375 frames (4:1 compression ratio).
\end{itemize}
\subsubsection{LLM}
We employ two families of LLM in our system: Qwen2.5-7B\footnote{https://huggingface.co/Qwen/Qwen2.5-7B-Instruct} with 7B parameters, and Gemma3-12B\footnote{https://huggingface.co/google/gemma-3-12b-it} with 12B parameters. Both LLMs have the capability to support an extensive number of languages.

\subsection{Data preparation}
The training set comprises around 1,500 hours of recordings in 11 languages: English \texttt{(en)}, French \texttt{(fr)}, German \texttt{(de)}, Italian \texttt{(it)}, Portuguese \texttt{(pt)}, Spanish \texttt{(es)}, Japanese \texttt{(jp)}, Korean \texttt{(ko)}, Russian \texttt{(ru)}, Thai \texttt{(th)}, Vietnamese \texttt{(vi)}. In English, there are 5 smaller subclasses: American, British, Filipino, Australian, and Indian. Each recording is a monolingual two-speaker conversation on random topics. 

To be compatible with pre-trained Whisper speech encoders, we segment each recording into 30-second segments with an overlapping section of 15 seconds. In total, we achieve around 2,300 hours of 30-second utterances for training. The challenge also provides a development set with the same settings as the training set, with approximately 4 hours of recordings for each language.

\subsection{Training details}
All training stages utilize Flash Attention 2 \cite{dao2023flashattention} for memory-efficient attention computation across both encoder and decoder components. All stages are trained using a learning rate of 3e-5 with a Cosine warmup ratio of $0.05$, optimized by AdamW \cite{loshchilov2017decoupled} with a weight decay of 1e-5. For augmentation, we only apply SpecAugment \cite{park19e_interspeech} to enhance the speech encoders' robustness. All models are trained on two NVIDIA A40 GPUs with DeepSpeed ZeRO-2 for efficient parallelization.

All models are evaluated with the Word Error Rate (WER\%). For Korean, Japanese, and Thai, we add a space between every character and calculate the Character Error Rate (CER\%). We use the \texttt{meeteval}\footnote{https://github.com/fgnt/meeteval} toolkit for evaluation, similar to the baseline implementation.

\subsubsection{Whisper-only system}
We fine-tune the Whisper large-v3 on 2,300 hours of the training set for 10 epochs. 
The fine-tuned Whisper implies Whisper large-v3 with the implementation details mentioned above in this paper, unless specified otherwise.
\subsubsection{Whisper and Qwen2.5}
We use the fine-tuned Whisper and train the system in the 3-stage manner as discussed in Section \ref{sec:system}. We use LoRA with an alpha value of $32$ to fine-tune the Qwen2.5-7B version with precision of 16 bits. The projector used is Projector 5.
\subsubsection{Whisper and Gemma3}
We also use the fine-tuned Whisper and train the system in the 3-stage manner as discussed in Section \ref{sec:system}. Note that in stage 2 for Gemma3, we continue to train the speech encoder along with the Projector 4 to achieve better feature alignment. We also use LoRA with an alpha of $32$ to fine-tune the Gemma3-12B version, with precision of 4 bits. 

\section{Experimental Results}
\subsection{Main results}
The main results are illustrated in Table \ref{tab:main-res}. In relative, our proposed systems outperform the baseline by 7.78\% and 17.55\% for Whisper+Qwen2.5-7B and Whisper+Gemma3-12B respectively. The integration of Gemma3 helps to reduce the CER/WER significantly, with an absolute reduction of 1.97\% compared to using Qwen2.5-7B as the language model. 

\begin{table}[]
\centering
\caption{Average CER/WER (\%) results on development and evaluation set}
\label{tab:main-res}
\resizebox{\columnwidth}{!}{%
\begin{tabular}{|l|rr|}
\hline
\textbf{Model} & \multicolumn{1}{l}{\textbf{Dev WER/CER (\%)}} & \multicolumn{1}{l|}{\textbf{Eval WER/CER (\%)}} \\ \hline\hline
Baseline           & -     &  20.17    \\
Top 1              & -     & 8.88     \\
Whisper+Qwen2.5-7B & 21.31     & 18.60 \\
Whisper+Gemma3-12B & 20.68 & 16.63 \\ \hline
\end{tabular}%
}
\end{table}

\begin{table*}[h!]
    \centering
    \caption{WER/CER (\%) for each language on the development set of the baseline systems and our models. \textbf{Bold} indicates the best result overall (row-wise), and \underline{underline} indicates the best result of every model for each language group.}
    \resizebox{\linewidth}{!}{%
    {\LARGE
    \begin{tabular}{lr|rr|rr|r|rrrr}
    \toprule 
    \textbf{Language} &
    \multicolumn{3}{c|}{\textbf{Organizer Baseline}} &
    \multicolumn{3}{c|}{\textbf{Our Baseline}} &
    \multicolumn{3}{c}{\textbf{Our SpeechLLMs}} \\
    \textbf{} & \textbf{\shortstack{Baseline\\LargeV3}} & \textbf{\shortstack{Baseline\\-Qwen}} & \textbf{\shortstack{Baseline\\-Llama}} & \textbf{\shortstack{LargeV3-I \\+ Qwen2.5-7B \\ (EC)}} & \textbf{\shortstack{Phi-4\\-multimodal\\0-shot}} & \textbf{LargeV3-I} & \textbf{\shortstack{Qwen2.5-7B\\16bit-III}} & \textbf{\shortstack{Gemma3-12B\\4bit-II}} & \textbf{\shortstack{Gemma3-12B\\4bit-III}} \\
    \midrule
    \textbf{English} \\
    English-American & 14.14 & \textbf{13.83} & 16.87 & 28.80 & 21.56 & 20.17 & \underline{20.25} & 21.19 & 20.50 \\
    English-Australian & 11.72 & \underline{11.19} & 13.32 & 19.62 & 16.83 & \textbf{9.68} & \underline{12.50} & 12.69 & 12.97 \\
    English-British & \textbf{10.08} & 11.00 & \underline{10.97} & 21.62 & 18.97 & 13.26 & \underline{15.02} & 15.70 & 15.64 \\
    English-Filipino & \underline{9.20} & \textbf{8.06} & 8.26 & 40.23 & 24.28 & 9.16 & \underline{10.25} & 11.21 & 10.28 \\
    English-Indian & \textbf{13.96} & 16.87 & \underline{15.67} & 21.19 & 19.63 & 15.66 & 16.34 & \underline{15.70} & 16.20 \\
    \midrule
    \rowcolors{2}{white}{gray!20}
    \textbf{European} \\
    French & 28.14 & \textbf{25.69} & 26.43 & 41.70 & 44.63 & 27.78 & 32.28 & 32.24 & \underline{31.15} \\
    German & \textbf{20.72} & 33.95 & \underline{32.37} & 40.78 & 40.13 & 26.73 & 32.90 & 33.85 & \underline{31.72} \\
    Italian & \textbf{17.92} & \underline{23.47} & 24.15 & 28.12 & 18.59 & 18.40 & 21.61 & 22.53 & \underline{20.54} \\
    Spanish & \underline{\textbf{12.27}} & \underline{14.31} & 16.41 & 21.47 & 16.49 &  12.84 & 15.38 & 15.97 & \underline{15.37} \\
    Portuguese & \textbf{21.23} & 34.02 & \underline{33.91} & 32.91 & 44.23 & 25.33 & 36.29 & \underline{32.70} & 31.09 \\
    Russian & 17.67 & \underline{18.25} & 19.07 & 23.79 & 608.22 & \textbf{14.51} & 18.18 & \underline{18.15} & 19.51 \\
    \midrule
    \rowcolors{2}{white}{gray!20}
    \textbf{East Asia} \\
    Japanese & \textbf{21.64} & 34.74 & \underline{33.82} & 38.88 & 55.48 & 23.64 & \underline{25.66} & 28.37 & 26.81 \\
    Korean & \underline{\textbf{13.80}} & \underline{20.77} & 22.56 & 39.63 & 420.00 & 16.53 & \underline{20.55} & 20.88 & 21.14 \\
    \midrule
    \rowcolors{2}{white}{gray!20}
    \textbf{South East Asia} \\
    Thai & \underline{14.49} & 21.67 & \underline{19.62} & 41.06 & 139.11 & \textbf{10.78} & 18.61 & 14.79 & \underline{13.25} \\
    Vietnamese & 27.16 & \underline{21.50} & 22.92 & 29.54 & 130.08 & \textbf{20.64} & \underline{23.76} & 25.34 & 24.00 \\
    \midrule
    \textbf{Avg.} & \textbf{16.94} & \underline{20.62} & 21.09 & 31.29 & 107.88 & 17.67 & 21.31 & 21.42 & \underline{20.68} \\
    \bottomrule
    \end{tabular}
    }
    }
    \label{tab:each-lang}
\end{table*}

\subsection{Ablation studies}

In this section, we provide in-depth results in each language on the development set in Table \ref{tab:each-lang}. We also divide the languages by group to see which language does every model perform best in each group. We compare our proposed systems with 4 baselines: (i) the baseline vanilla Whisper, which involves fine-tuning a single Whisper-large-v3 and use that as the transcriber \textbf{(Baseline LargeV3)}; (ii) the vanilla Whisper as the speech encoder and Qwen2.5-7B as a language model fine-tuned with LoRA \textbf{(Baseline-Qwen)}; (iii) the vanilla Whisper and Llama3.1-8B \cite{gao2023llama} fine-tuned with LoRA \textbf{(Baseline-Llama}; and (iv) Phi-4 \cite{abdin2024phi} - a multimodal LLM, transcribing in a zero-shot manner \textbf{(Phi-4-multimodal-0-shot)}. Note that Phi-4 was not pre-trained on Russian, Korean, Thai, and Vietnamese among the evaluated languages. We use the instruction-fine-tuned version, Phi-4-Instruct for inference. Our proposed systems for comparison include the following:
\begin{itemize}
    \item \textbf{LargeV3-I.} The Whisper-large-v3 fine-tuned on the provided training data.
    \item \textbf{Qwen2.5-7B-16bit-III.} The fine-tuned Whisper along with the Qwen2.5-7B fine-tuned with LoRA to stage 3.
    \item \textbf{Gemma3-12B-4bit-II.} The fine-tuned Whisper along with Gemma3-12B fine-tuned in LoRA to stage 2.
    \item \textbf{Gemma3-12B-4bit-III.} The fine-tuned Whisper along with Gemma3-12B fine-tuned in LoRA to stage 3.
\end{itemize}

We can first see that Phi-4 Instruct, a public LLM baseline, performs worse than all other baselines and custom models, with an average WER/CER of 107.88\%. In contrast, the average of other baselines ranges from 16.94\% (Baseline LargeV3) to 21.09\% (Baseline-Llama), indicating much more stable and realistic performance.

A clear trend in the table is that direct integration of Whisper with LLMs like Qwen2.5 and Llama3.1 leads to performance degradation compared to vanilla Whisper. For example, for almost every language, Baseline-Qwen and Baseline-Llama yield higher WER/CER than vanilla Whisper. This suggests that naive fusion with large language models leads to degraded recognition performance.

While not universally superior, our LargeV3-I significantly improves over Baseline-LargeV3 in several languages. For example, it reduces error rates in English-Australian (11.72\% to 9.68\%), English-Filipino (9.20\% to 9.16\%), French (28.14\% to 27.78\%), Russian (17.67\% to 14.51\%), Thai (14.49\% to 10.78\%), and Vietnamese (27.16\% to 20.64\%).

When comparing our LargeV3-I + Gemma-12B-4bit-III model with the two baseline fused models (Baseline-Qwen and Baseline-Llama), it performs better on nearly every language, achieving a relative error reduction of 1.95\% over Baseline-Llama, while slightly underperforming Baseline-Qwen with a marginal increase of 0.29\%. Overall, both our Qwen2.5-7B-16bit-III and Gemma3-12B-4bit-III configurations outperform the baselines in the East Asia and Southeast Asia language groups, but lag behind in English and European languages.

We also added a LargeV3-I + Qwen2.5-7B model for Error Correction (EC) as a cascaded version of SpeechLLM, where the LLM will fix the transcription output by Whisper. While it shows promising results, it actually degrades performance compared to the original LargeV3-I output (increasing the error from 17.67\% to 31.29\%) and still lags behind the Qwen2.5-7B-16bit-III model (21.31\%). This showcases the effectiveness of end-to-end optimization. Note that this experiment is for ablation only, since the challenge does not permit the use of LLM as an supplementary EC.

\section{Conclusions}
We present a system for the MLC-SLM Challenge 2025 that effectively combines state-of-the-art speech encoder and language modeling components. Our three-stage training methodology and modular architecture enable flexible experimentation and optimization of different system configurations.

The experimental results demonstrate the effectiveness of our approach, with the Gemma3-12B configuration achieving an average CER/WER of 16.63\% on the final private evaluation set and securing a rank of 15 out of 25. The system successfully handles multilingual speech recognition across 15 different accents and languages, showcasing robust performance in diverse linguistic contexts.


Future work will focus on exploring additional language model architectures, investigating more sophisticated projector designs, and extending the approach to handle even more diverse linguistic variations and acoustic conditions.

\section{Acknowledgements}
The authors would like to thank the MLC-SLM Challenge 2025 organizers for providing the dataset and evaluation framework. We also acknowledge the computational resources provided by the Institute for Infocomm Research (I$^2$R), A$^*$STAR, Singapore, which made this research possible.

\bibliographystyle{IEEEtran}

\bibliography{mybib}

\end{document}